\documentclass[times,twocolumn,final,authoryear]{elsarticle}

\usepackage{prletters}
\usepackage{framed,multirow}

\usepackage{amssymb}
\usepackage{latexsym}

\usepackage{amsmath}
\usepackage{mathtools}

\usepackage{url}
\usepackage{xcolor}
\definecolor{newcolor}{rgb}{.8,.349,.1}

\journal{Pattern Recognition Letters}

\begin{document}

\begin{frontmatter}

\title{Approximate spectral clustering with eigenvector selection and self-tuned $k$}

\author{Mashaan \snm{Alshammari}\corref{cor1}}
\cortext[cor1]{Corresponding author}
\ead{mals6571@uni.sydney.edu.au}

\author{Masahiro \snm{Takatsuka}}
\ead{masa.takatsuka@sydney.edu.au}

\address{School of Information Technologies, The University of Sydney, NSW 2006 Australia}

\received{1 May 2013}
\finalform{10 May 2013}
\accepted{13 May 2013}
\availableonline{15 May 2013}
\communicated{S. Sarkar}

\begin{abstract}
The recently emerged spectral clustering surpasses conventional clustering methods by detecting clusters of any shape without the convexity assumption. Unfortunately, with a computational complexity of $O(n^3)$, it was infeasible for multiple real applications, where $n$ could be large. This stimulates researchers to propose the approximate spectral clustering (ASC). However, most of ASC methods assumed that the number of clusters $k$ was known. In practice, manual setting of $k$ could be subjective or time consuming. The proposed algorithm has two relevance metrics for estimating $k$ in two vital steps of ASC. One for selecting the eigenvectors spanning the embedding space, and the other to discover the number of clusters in that space. The algorithm used a growing neural gas (GNG) approximation, GNG is superior in preserving input data topology. The experimental setup demonstrates the efficiency of the proposed algorithm and its ability to compete with similar methods where $k$ was set manually.
\end{abstract}


\end{frontmatter}



\section{Introduction}
\label{Introduction}


Spectral clustering (\cite{RN261,RN233,RN234}) emerged to be an effective learning tool that attempts to capture non-convex similarities in input data. Unlike spherical clustering algorithms, spectral clustering does not assume compactness of clusters, instead it is driven by the connectivity between data points. This enables spectral clustering to uncover more complex shaped clusters leading to efficient clustering. It was used in many applications like: image segmentation (\cite{RN261,RN228}), object localization (\cite{RN286}), and community networks (\cite{RN292}). Unfortunately, its computational cost prevented it from expanding to more practical problems, since its core component is decomposing the graph Laplacian $L$ of size $n \times n$, leaving the algorithm with a complexity of $O(n^3)$ (\cite{RN232}).


Due to its ability of providing high quality clustering, spectral clustering computational demands were well studied in the literature. The most intuitive solution is to sample representatives from the dataset to perform spectral clustering then generalize the results. Clearly, these methods ignore data points dependencies while performing sampling, which could lead to a loss of small clusters. Therefore, sampling was replaced by vector quantization to select $m$ representatives (\cite{RN232}) where $m \ll n$. This approach accumulates the feature space dependencies into a set of representatives to perform spectral clustering on.

Most of approximate spectral clustering methods assumed that the number of clusters $k$ was known beforehand. This is a strong assumption giving that tuning $k$ is not straightforward for many real applications (e.g., image segmentation \cite{RN294}). The original work by (\cite{RN233}) indicates two uses of the parameter $k$. The first use related to selecting $k$ eigenvectors corresponding to minimum eigenvalues of the normalized graph Laplacian $L$, those eigenvectors represent the embedding space $\mathbb{R}^{n \times k}$. Secondly, $k$ was used by $k$-means to separate data points.
\begin{figure*}
	\centering
	\includegraphics[width=0.9\textwidth,height=20cm,keepaspectratio]{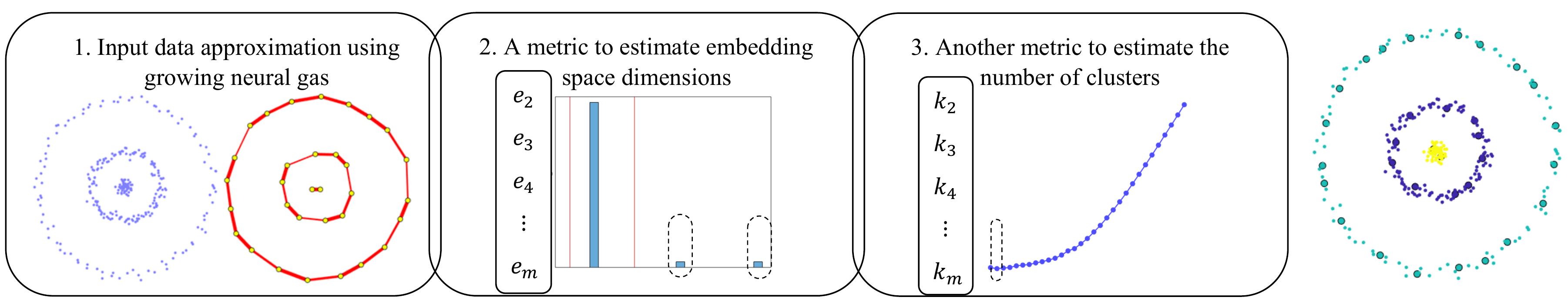}	
	\caption{Illustration of the proposed approach.}
	\label{Fig:Flowchart}
\end{figure*}

We attempt to estimate the parameter $k$ as well as reducing the complexity of spectral clustering. Initially, a growing neural gas (\cite{RN270}) is trained on the feature space to produce representative neurons. Then, the neurons' affinity matrix was constructed to decompose the graph Laplacian. The first need of $k$ was removed by selecting the eigenvectors that maximize a relevance metric based on separation between the graph nodes and explained variance. The final step performed by clustering neurons in the embedding space using $k$-means, where the number of clusters was estimated based on another metric. Although spectral clustering could be used in a range of applications, our experiments were mainly focused on image segmentation problems for two reasons. First, the studies that introduced these datasets used a spectral clustering benchmark with a manual $k$. Second, the effect of automating $k$ could be easily visualized in image segmentation problems. The proposed method showed a competitive performance to methods where $k$ was manually tuned.

This work is organized as follows: in section \ref{RelatedWork} we review efforts related to approximations of spectral clustering; sections \ref{ProposedApproach} and \ref{Experiments} introduce our proposed solution followed by experimental results.
\section{Related Work}
\label{RelatedWork}

Given a graph $G(V,E)$ connecting data points and the affinity matrix $A$, spectral clustering attempts to relax the normalized cut problem (Ncut) introduced by \cite{RN261}. \cite{RN233} proposed a symmetric graph Laplacian as $L_{sym}=I-D^{-\frac{1}{2}}AD^{-\frac{1}{2}}$. Then the points $\{x_1, x_2, \cdots, x_n\}$ mapped into the embedding space $\mathbb{R}^{n\times k}$ spanned by $k$ smallest eigenvectors. In $\mathbb{R}^{n\times k}$, points form convex clusters that could be detected by running $k$-means to avoid iterative bipartitioning. This algorithm highlighted two uses of the parameter $k$. First, it is used to select the top $k$ eigenvectors of the graph Laplacian $L$, then used as an input for $k$-means for clustering. Efforts in the literature for automating $k$ could be classified based on their source for evaluation, either eigenvalues or eigenvectors.

One way to discover $k$ is to count the number of eigenvalues with multiplicity zero. Alternatively, one could look for the largest gap between $k$ and $k+1$ in the eigenvalues plot, this method is known as eigengap. Ideally, the gap between $k$ and $k+1$ is large (\cite{RN234}). However, these techniques need a clearly separated data (\cite{RN237}). \cite{RN168} proposed a soft threshold to locate the eigengap. They used a parameter $\tau$ to penalize small eigenvalues that are less important for clustering. Although $\tau$ has fixed limits (i.e., $0<\tau<1$), it still needs a carful tuning.

The use of eigenvectors to estimate $k$ was initiated by (\cite{RN237}) where they look for an optimal rotation $\hat X$ of the matrix of eigenvectors $X$. The method starts by recovering the rotation of the first two eigenvectors then iteratively add one eigenvector to be rotated. A cost function is computed with each iteration, and the optimal set is the one that yields the minimal cost. One deficiency of this method is the need for the parameter $k_{max}$, if not set, the method could have to rotate a $\mathbb{R}^{m \times m}$ space which could be costly
. This approach was improved by (\cite{RN288}) where they used a computationally efficient alignment cost. Also, they emphasized on the effectiveness the initialization scheme for better optimization. Although these methods are well formulated with sophisticated optimization, they could be computationally inefficient. They require high dimensional optimization space and multiple initializations.

Eigenvectors of $L$ could be evaluated individually to avoid the need for high dimensional optimization space. \cite{RN265} proposed an eigenvector evaluation metric $R_{e_k}$ based on the distribution of the points, whether it is unimodal and multimodal. Ultimately, eigenvectors with low discrimination power ended up with low $R_{e_k}$ scores. The number of clusters in the embedding space was estimated based on the lowest Bayesian information criterion (BIC). Another effort for estimating $k$ was introduced by \cite{RN264}, where they ranked the eigenvectors based on the entropy caused by the absence of that eigenvector. Nevertheless, to obtain the entropy score, one might need to examine different combinations of eigenvectors. \cite{RN293} formulated a cost function to evaluate eigenvectors based on intra-cluster compactness and inter-cluster separation. Starting from $k=2$, eigenvectors were evaluated and the set with minimum score will be returned. All aforementioned methods require a parameter $k_{max}$ which is safely larger than the true number of clusters $k_{true}$. In addition, they have not used the eigenvalues of $L$ which they could possess important information.

The need for the parameter $k$ could be eased by avoiding the method of \cite{RN233} (known as $k$-way approach). Alternatively, the graph would go through iterations each of which uses a single eigenvector to bipartition the graph. This process of iterative spectral clustering was used by \cite{RN290} to partition the graph recursively. A cluster is established if it cannot find a gap that satisfies the minimum tolerance. It also was used by \cite{RN292} to detect community networks. They define a ``critical edge’’ to highlight the most significant bipartition. Iterative spectral clustering was used by \cite{RN286} for object localization. The main difficulty of iterative spectral clustering is how to specify the stopping criteria. Also, it processes the eigenvectors independently \cite{RN261}.

The deficiencies of previous efforts to automate $k$ could be summarized in three points. First, they use one source of information, either eigenvalues or eigenvectors. Second, they estimated one value for the number of eigenvectors and the number of clusters, however, these two values are not necessarily equal. Third, they used iterative clustering that is less efficient than using $k$ eigenvectors simultaneously. We attempted to avoid these shortcomings by proposing two evaluation metrics that utilize both eigenvalues and eigenvectors.

\section{Proposed Approach}
\label{ProposedApproach}

The proposed method provides an informed measure to estimate the appropriate $k$. It starts by mapping input data into a growing neural gas (GNG). The spectral clustering continues by decomposing the graph Laplacian $L$. The obtained eigenvectors were examined against a relevance metric to select the ones that provides best separation of graph nodes penalized by their eigenvalues. In the embedding space $\mathbb{R}^{m \times k}$, the value of $k$ was estimated by another relevance metric that measures the separation of clusters and the accumulated sum of eigenvalues.


\subsection{Growing Neural Gas}
Growing neural gas (GNG) was proposed by \cite{RN270}, it was an improvement over self-organizing map (SOM) 
and neural gas (NG). 
GNG starts by introducing a random point $x_i$ to the competing neurons. The winning neuron is the nearest and called the best matching unit $w_{b}$:
\begin{equation}
	\lVert x_i-w_{b} \rVert = \min_{i}\{ \lVert x_i-w_i \rVert \}
	\label{Eq-SOM-001}
\end{equation}
\noindent
Then GNG computes the error of $w_{b}$ using:
\begin{equation}
	error(w_{b},t+1) = error(w_{b},t) + \lVert w_{b}-x_i \rVert^2
	\label{Eq-GNG-001}
\end{equation}
\noindent
The new positions of $w_{b}$ and its topological neighbors are computed as per the following adaption rules:
\begin{equation}
	\label{Eq-GNG-002}
	\begin{aligned}
		w_{b}(t+1) = \epsilon_b (x_i - w_{b}) \\
		w_{d}(t+1) = \epsilon_k (x_i - w_{d})
	\end{aligned}	
\end{equation}

$d$ represents all direct topological neighbors of $w_{b}$, whereas $\epsilon_b$ and $\epsilon_d$ determine the amount of change. GNG introduces a new neuron to the map if the current iteration is a multiple of some parameter $l$. It first finds the neuron with the maximum accumulated error $w_q$, then determines its neighbor with the largest error $w_f$. The new neuron is inserted between $w_q$ and $w_f$ and connects to both of them. The training stops if quantization error is stable.

\subsection{Approximating Data in Feature Space}
To perform spectral clustering on large input data, a preprocessing step using growing neural gas was deployed to minimize the input data. GNG was selected over the original preprocessing of $k$-means (\cite{RN232}) for two reasons. First, unlike $k$-means that places representatives and let the similarity measure connects them, GNG places representatives and connects them with edges. This leaves the similarity measure with an easy task of only weighing those edges produced by GNG, which is usually a sparse graph. Second, GNG uses the competitive Hebbian rule to draw edges, which produces a graph called \textit{``induced Delaunay triangulation''}. This graph forms a perfectly topology preserving map of input data (see Theorem 2 in \cite{RN269} and the discussion therein).

\begin{figure}
	\includegraphics[width=0.48\textwidth,height=20cm,keepaspectratio]{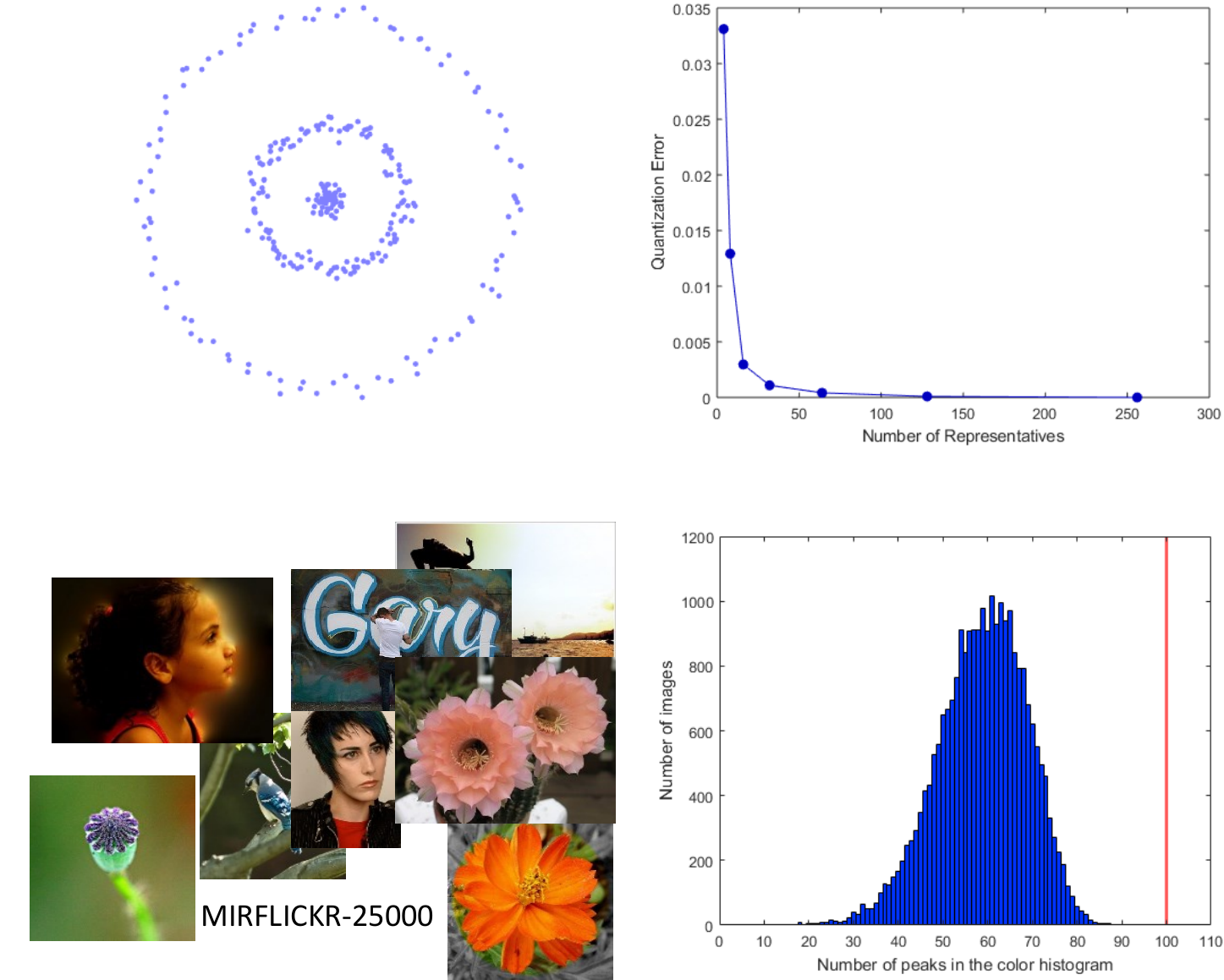}
	
	\caption{(Top) Synthetic data quantization error for $m \in \{4,8,16,32,64,128,256\}$, showing that $m=32$ is an elbow point. (Bottom) Distribution of the number of peaks in the color histogram for 25000 images retrieved from MIRFLICKR-25000. The red vertical line is the size of GNG (best viewed in color).}
	\label{Fig:SelectM}
\end{figure}

The most influential parameter of GNG is its size ($m$). In case of images, ideally, we would like to capture all color patterns with less number of neurons. Therefore, setting the lower bound for GNG size was very critical. To uncover color patterns, we examined color histogram peaks in 25000 images retrieved from MIRFLICKR-25000 (\cite{RN249}). The distribution of color peaks illustrated in Fig. \ref{Fig:SelectM}, in which it is observable that setting $m$ in GNG to $100$ neurons is sufficient to capture all color patterns. In case of synthetic data, a range of $m$ values were examined using $k$-means++ algorithm. $m$ was set as the number that represents an elbow point in the quantization error curve (Fig. \ref{Fig:SelectM}).

\subsection{Performing Spectral Clustering}
When $m$ neurons are trained to approximate $n$ points ($m \ll n$), they became ready to be processed by spectral clustering. The first step is to construct the affinity matrix $A=\{A_{ij}\}_{i,j=1}^m$, where $A_{ij}$ denotes the similarity between $w_i$ and $w_j$. Commonly, $A$ is constructed by a kernel with a global scale $\sigma$.
In spite of its popularity, global $\sigma$ processed all data points equally regardless of their status in the feature space. Therefore, it could be challenged when input data contains different local statistics (\cite{RN237}). A more reasonable selection of the scaling parameter is to set it locally as:

\begin{equation}
	A_{ij} = exp \bigg(
	\frac{-d^2(w_i,w_j)}{\sigma_i \sigma_j}
	\bigg) \quad \text{, where } (i,j) \in E_{GNG}
	\label{Eq-ScaleLocal1}
\end{equation}
\noindent
The local scale $\sigma_i$ could be set as:

\begin{equation}
	\sigma_i = d(w_i,w_K)
	\label{Eq-ScaleLocal2}
\end{equation}

$w_K$ is the $K^{\text{th}}$ neighbor of $w_i$. In \cite{RN237}, it was set as $K=7$, however, in our case it was set as $K=1$, that is the direct neighbor of $w_i$. This was consequent to the reduction performed by GNG in the preprocessing. The degree matrix $D$ is defined as $D_{ii}=\sum_{j} A_{ji}$. The diagonal in $D$ denotes the degree for all $\{w_i\}_{i=1}^m$. Then, the normalized graph Laplacian is computed as $L_{sym}=D^{-\frac{1}{2}}LD^{-\frac{1}{2}}=I-D^{-\frac{1}{2}}AD^{-\frac{1}{2}}$ (\cite{RN234}).

\subsection{Evaluation of Graph Laplacian Eigenvectors}
Decomposing $L_{sym}$ produces $m$ nonnegative real-valued eigenvalues $0=\lambda_1 \le\cdots\le \lambda_m$. The graph nodes are separable in the space $\mathbb{R}^{m \times k}$ spanned by the eigenvectors corresponding to $k$ smallest eigenvalues. One way to uncover $k$ is to count the eigenvalues of multiplicity 0 (\cite{RN234}). However, $k$ is not always clear in eigenvalues and relying on this to uncover the number of clusters might not hold in case of noise (\cite{RN237}). Another way to estimate the number of dimensions in $\mathbb{R}^{m \times k}$ is to recover the rotation that best aligns the neurons to a block-diagonal matrix $L$ (\cite{RN237}). The combination of eigenvectors that best recover such an alignment is the optimal set to construct $\mathbb{R}^{m \times k}$. However, recovering the alignment becomes more expensive as we approach $m$, since an $m \times m$ space needs to be rotated. This entails the tuning of another parameter that is $k_{max}$. 

According to \cite{RN234}, the eigenvector corresponding to the second smallest eigenvalue provides a solution to the Normalized Cut (NCut) problem. Therefore, the first eigenvector indicates the connectivity of the graph and the second indicates the largest cut in the graph, then more eigenvectors are included if there are more clusters.
\begin{figure}
	\includegraphics[width=0.48\textwidth,height=20cm,keepaspectratio]{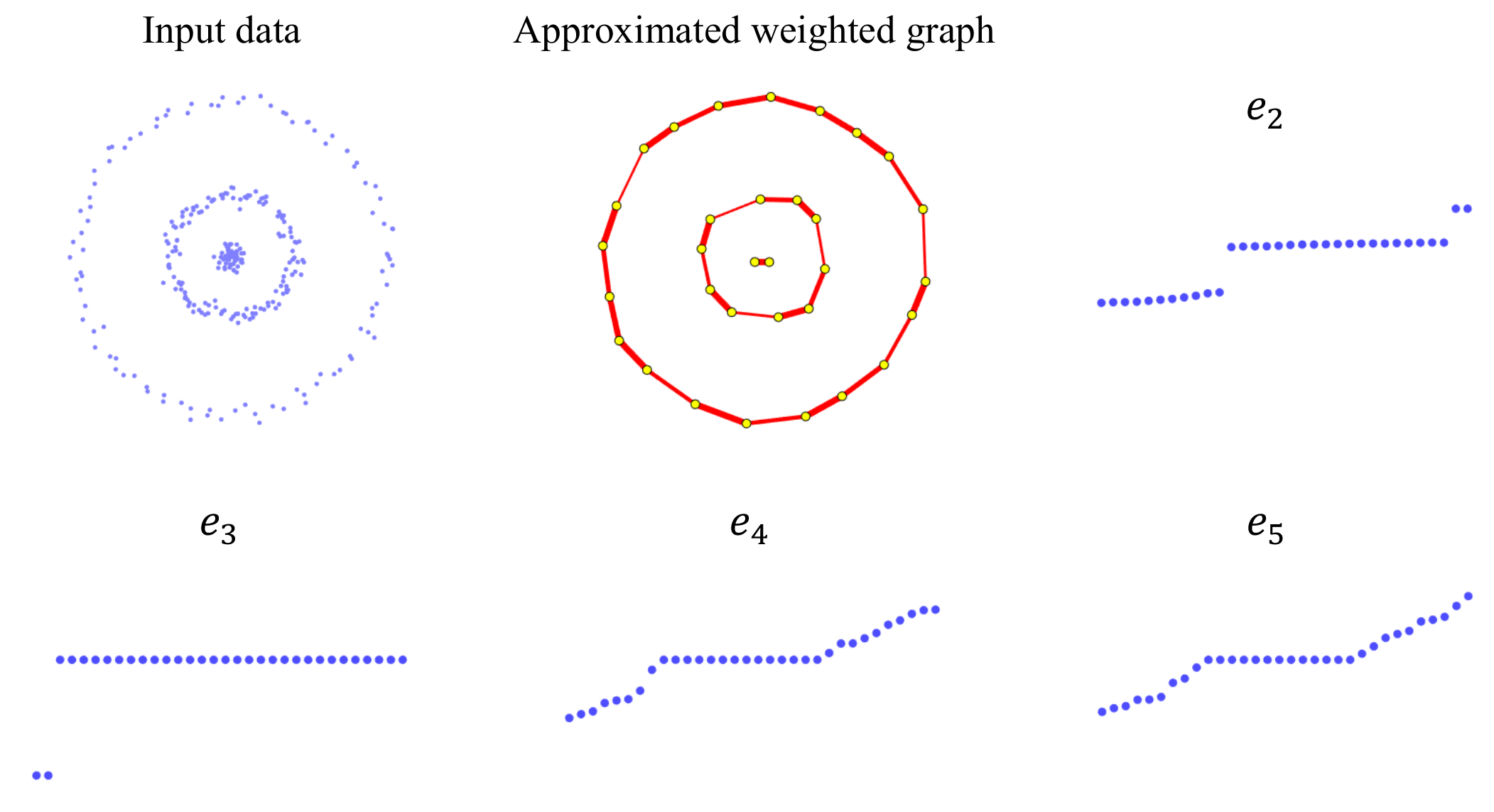}
	
	\caption{Top eigenvectors of the approximated graph for the synthetic data.}
	\label{Fig:3RingsV}
\end{figure}
Fig. \ref{Fig:3RingsV} shows an empirical validation on the usefulness of the second eigenvector $e_2$. Given that the true $k$ equals 3, the $e_2$ and $e_3$ eigenvectors are the most informative. In contrast, $e_4$, $e_5$, and $e_6$ contain no separation of graph nodes. Therefore, we define a relevance metric $R_{e_k}$ that measures the ability of separating the graph nodes into 2, 3, and 4 clusters. Additionally, this quantity was penalized by the eigenvalue corresponding to that eigenvector, to promote eigenvectors with smaller eigenvalues:

\begin{equation}
	R_{e_k} = \frac{\sum_{c=2}^{4} DBI_c(e_k)}{\lambda_i} \quad \text{, } 1 \leq i \leq m
	\label{Eq-Cost102}
\end{equation}
\noindent
$DBI_c$ is the Davies--Bouldin index value defined as: 

\begin{equation}
	DBI_c(e_k) =
	\frac{1}{c} \sum_{i=1}^{c} \max_{i\neq j} \bigg\{
	\frac{S_c(Q_i)+S_c(Q_j)}{d_{ce}(Q_i,Q_j)}
	\bigg\}
	\label{Eq-DBI}
\end{equation}

Given one dimensional data $\{w_1,w_2,\cdots,w_n\}$ in $e_k$, it could be clustered into $\{Q_1,Q_2,\cdots,Q_c\}$ clusters where $c \in \{2,3,4\}$. $S_c(Q_i)$ is within-cluster distances in cluster $i$, and $d_{ce}(Q_i,Q_j)$ is the distance between clusters $i$ and $j$.

\begin{figure}
	\includegraphics[width=0.48\textwidth,height=20cm,keepaspectratio]{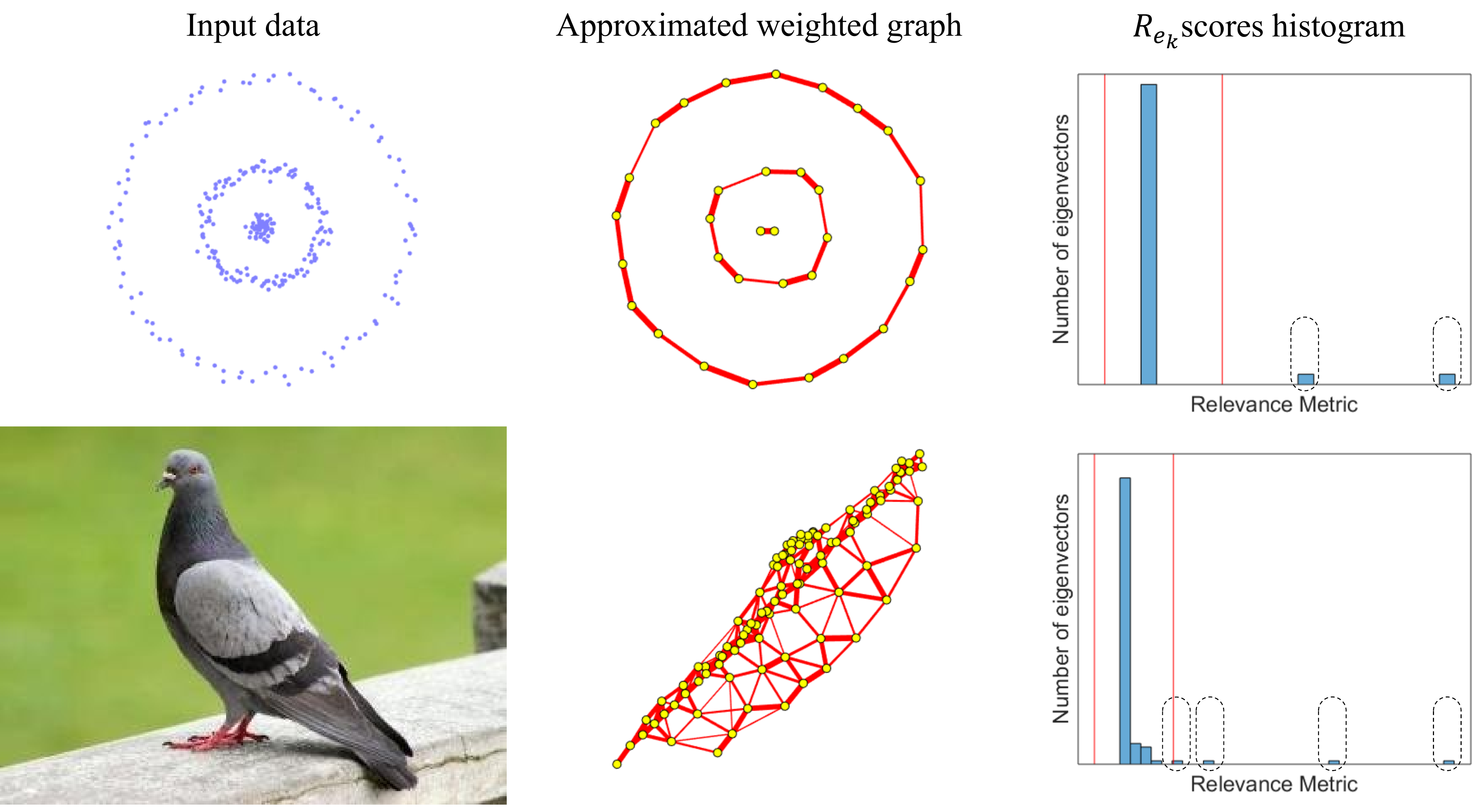}
	
	\caption{Input data approximated using GNG, and the weights obtained through local $\sigma$. The $R_{e_k}$ histograms suggest that two eigenvectors are sufficient to cluster synthetic data and four eigenvectors to segment the image. The red vertical lines represent the interval $[\mu \pm \sigma]$ (best viewed in color).}
	\label{Fig:RelevanceScore}
\end{figure}

$R_{e_k}$ enables us to attach a score with every eigenvector for evaluation in order to select the most informative eigenvectors.
By definition, an informative eigenvector has a substantially large $R_{e_k}$ due to its small eigenvalue and large DBI score. Consequently, that eigenvector should standout from the remaining eigenvectors which are close to the mean score ($\mu$) of $R_{e_k}$. Therefore, the selected eigenvectors are the ones that fall outside the interval $[\mu \pm \sigma]$, where $\mu$ is the mean $R_{e_k}$ score and $\sigma$ is the standard deviation of $R_{e_k}$ scores. The bin size of this histogram was set as per Freedman-–Diaconis rule that is defined as $2Rm^{-1/3}$, where $R$ is the inter--quartile range. 
The qualified eigenvectors from $R_{e_k}$ histogram, constitute the matrix $X \in \mathbb{R}^{m \times k}$.

\begin{figure}
	\includegraphics[width=0.48\textwidth,height=20cm,keepaspectratio]{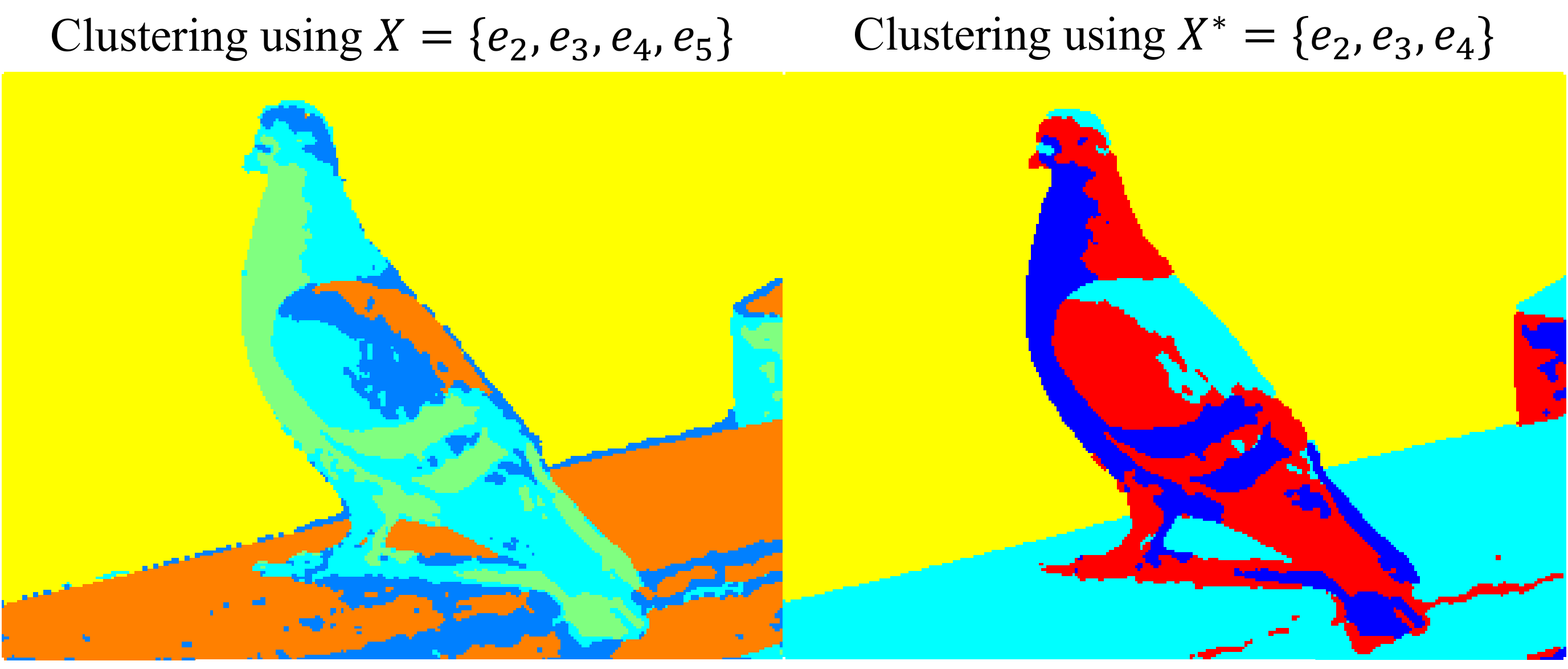}
	
	\caption{clustering results obtained through $X$ and $X^*$.
	}
	\label{Fig:ExplainVar}
\end{figure}

On synthetic data, $R_{e_k}$ performs efficiently to highlight the informative eigenvectors, since the graph is usually disconnected and the data is clearly separable. However, this is not the case in realistic data (e.g., images). Such data contain interconnected components that might promote unnecessary eigenvectors. Therefore, certain precautions have to be taken to make sure that $X$ is discriminative for $k$-means to find the true clusters. In Fig. \ref{Fig:ExplainVar}, we segmented the pigeon image using the eigenvectors qualified from $R_{e_k}$ histogram shown in Fig. \ref{Fig:RelevanceScore} (i.e., $e_2$, $e_3$, $e_4$, and $e_5$). Interestingly, we could achieve more polished segmentation if we dropped the last eigenvector. This is justified by the amount of variance explained by the eigenvectors. $e_2$, $e_3$, and $e_4$ accumulate almost $80\%$ of the variance. Hence, to ensure that $X$ remains discriminative, we performed a principle component analysis (PCA) on the eigenvectors qualified from $R_{e_k}$ histogram, and kept the ones that represent $80\%$ of the variance. We called the obtained matrix $X^* \in \mathbb{R}^{m \times k}$.

\subsection{Clustering in the Embedding Space $\mathbb{R}^{m \times k}$}
\begin{figure}
	\centering	
	\includegraphics[width=0.38\textwidth,height=20cm,keepaspectratio]{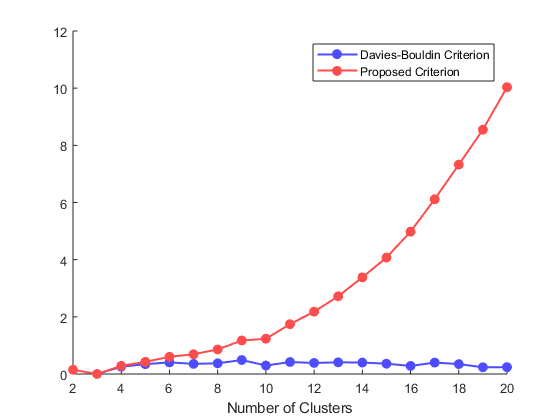}
	\caption{Testing a range of $k$ values to return the one with the lowest score (data in Fig. \ref{Fig:3RingsV}). With large $k$ values, DBI tends to give a good score due to better separation. But our proposed criterion penalizes large $k$ values and increases sharply to highlight the true $k$ ($k=3$) (best viewed in color).}
	\label{Fig:EmbedSpace002}
\end{figure}

The embedding space $\mathbb{R}^{m \times k}$ is spanned by the eigenvectors selected by $R_{e_k}$ histogram. In this space, the graph nodes form convex clusters. Therefore, they could be detected by $k$-means. One issue needs to be addressed is the number of clusters in the embedding space to run $k$-means. An intuitive way, to estimate $k$ is to measure Davies--Bouldin index over multiple $k$ values. However, DBI tends to favor large values of $k$ for better separation (see Fig. \ref{Fig:EmbedSpace002}). A better approach would be utilizing the graph Laplacian eigenvalues to penalize large values of $k$ that accumulate large sum of eigenvalues. Proposition 2 in (\cite{RN234}) states that \textit{``the multiplicity $k$ of the eigenvalue 0 of $L$ equals the number of connected components in the graph''}. Hence, a small sum of eigenvalues is preferable while examining different $k$ values. Simply, this metric looks for the value of $k$ that provides best separation of graph nodes and at the same time accumulates a small sum of eigenvalues.

\begin{equation}
	R_k = DBI_{k}(X^*) + \sum_{i=1}^{k}\lambda_i \quad \text{, } 2 \leq k \leq m
	\label{Eq-Cost202}
\end{equation}

\section{Experiments}
\label{Experiments}
Five methods were used to estimate the number of dimensions for the embedding space $\mathbb{R}^{m \times k}$. 1) eigengap method (\cite{RN234}), 2) eigenvectors alignment method (\cite{RN237}), 3) Low-Rank Representation metric used in (\cite{RN168}) with $\tau=0.08$, 4) $X$, where the number of dimensions was estimated via $R_{e_k}$ score, 5) $X^*$ is a refined version of $X$ to keep the eigenvectors that hold 80\% of the variance. To maintain a fair comparison amongst competing methods, the number of clusters in $\mathbb{R}^{m \times k}$ was estimated using the metric in equation \ref{Eq-Cost202}. All experiments were implemented in MATLAB 2017b and carried out on a Windows 10 machine with 3.40 GHz CPU and 8GB of memory.

\subsection{Synthetic data}
Table \ref{Tab:Ex-Toy} shows the clustering results of 100 runs for every method, where $m$ was selected as the elbow point in the quantization error curve. Some of used data was provided in supplementaries\footnote{\url{ http://lihi.eew.technion.ac.il/files/Demos/SelfTuningClustering.html }} of (\cite{RN237}).
\begin{table}
	\centering	
	\caption{Clustering results on synthetic datasets.
	}
	\label{Tab:Ex-Toy}
	
	\includegraphics[width=0.49\textwidth,height=20cm,keepaspectratio]{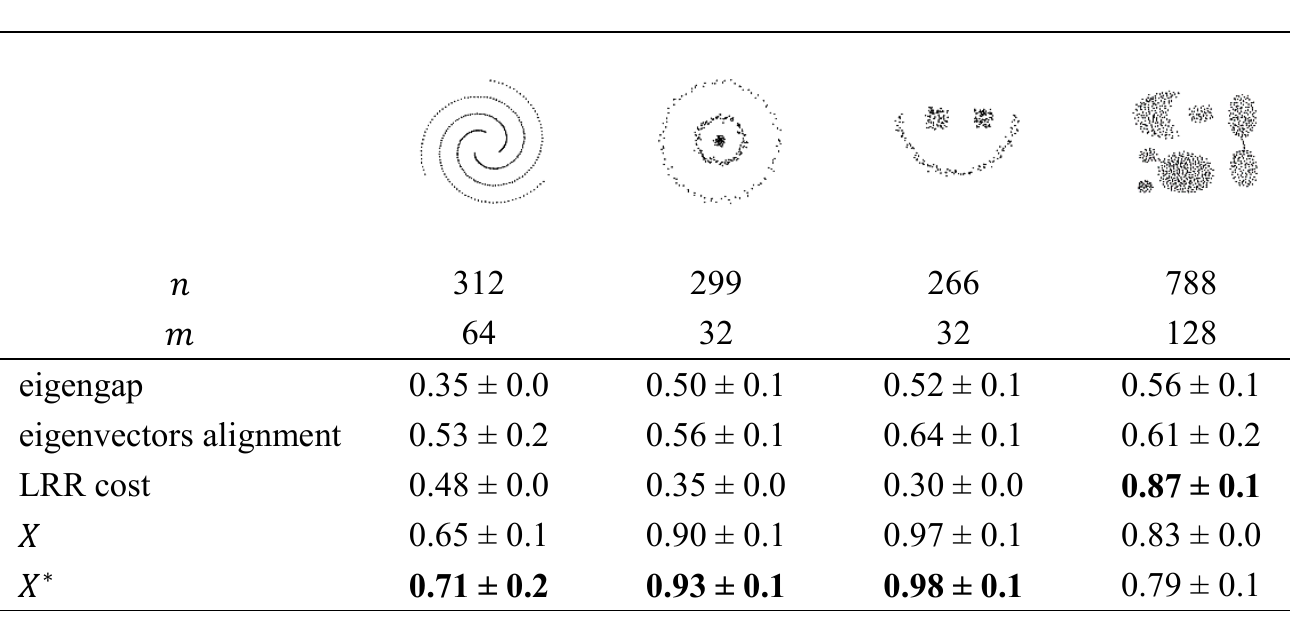}
\end{table}
The efficiency of the proposed method ($X^*$) was clearly demonstrated across the four datasets, particularly in second and third datasets where it was close to full mark. Eigengap and eigenvectors alignment did not perform as well as $X$ and $X^*$. Their low performances were due to large number of eigenvectors passed as dimensions for $\mathbb{R}^{m \times k}$, which confuses the number of clusters estimation in that space. The performance of LRR cost fluctuates across different datasets which highlights the large influence of the parameter $\tau$. 

\subsection{Real Images}
Weizmann segmentation evaluation dataset\footnote{\url{http://www.wisdom.weizmann.ac.il/~vision/Seg_Evaluation_DB/}} contains 100 RGB images, each of which has a single foreground object. Also, it provides three versions of human segmentation. The accuracy of the segmentation method was measured in terms of F-measure for the foreground class segmented by humans. 
This dataset was introduced by \cite{RN221}, where authors used the original version of spectral clustering (NCut) (\cite{RN261}) for benchmarking. That version used manual selection for the parameter $k$, hence, we used that score as a baseline to evaluate the competing methods in this study. The produced segmented image was post processed using $3 \times 3$ median filter to smooth small artifacts.

In Table \ref{Tab:Ex-Wiess}, the baseline score was $0.72$ for the spectral clustering where $k$ was manually set. The competing methods deviated from the baseline score by $-0.24$, $-0.19$, $-0.15$, $-0.11$, and $-0.08$. This observation demonstrates the efficiency of the proposed method compared to the baseline.
The first 3 methods tend to provide more eigenvectors for the embedding space $\mathbb{R}^{m \times k}$ which makes the task of detecting the number of clusters difficult. For $X$, the proposed metric was good, but we could achieve a better performance if only keep the eigenvectors that hold 80\% of the variance as in $X^*$.

\begin{table}
	\centering
	\caption{Results on Weizmann segmentation evaluation dataset. 
		The number in parentheses indicates the deviation from the baseline score.}
	\label{Tab:Ex-Wiess}
	\includegraphics[width=0.35\textwidth,height=20cm,keepaspectratio]{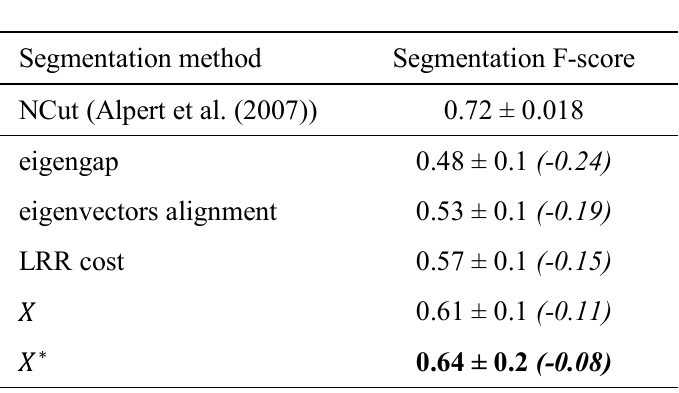}	
\end{table}

\begin{table}
	\centering
	\caption{Segmentation results on BSDS500. 
	The number in parentheses indicates the deviation from the baseline score.}
	\label{Tab:Ex-BSDS}
	\includegraphics[width=0.48\textwidth,height=20cm,keepaspectratio]{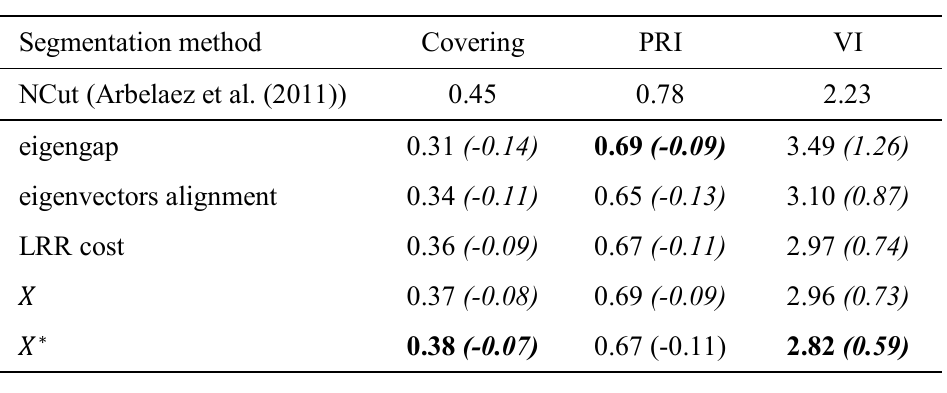}	
\end{table}

Berkeley Segmentation Data Set (BSDS500)\footnote{\url{https://www2.eecs.berkeley.edu/Research/Projects/CS/vision/grouping/resources}} is a more comprehensive dataset to evaluate segmentation methods. It contains 500 images alongside their human segmentation. It also provides three segmentation evaluation metrics: segmentation covering, probabilistic rand index (PRI), and variation of information (VI). A good segmentation would result in high covering, high PRI, and low VI scores. \cite{RN281} tested a version of spectral clustering on (BSDS500), it is called Multi Scale NCut (\cite{RN283}). The scores on that study were used as a baseline for the competing methods here. In that spectral clustering implementation, $k$ was set manually and the size of the affinity matrix $A$ was $n \times n$.

As illustrated in Table \ref{Tab:Ex-BSDS}, eigengap performed badly in terms of covering and VI, although it got a good PRI score. 
Struggling performances by the first three methods were due to the large number of eigenvectors passed for the embedding space. Better scores were achieved by $X$ and $X^*$ with a clear advantage for $X^*$. This emphasizes that keeping the eigenvectors that are accountable for 80\% of the variance would produce better segmentation than using all eigenvectors. Comparing $X^*$ to NCut scores reported in \cite{RN281}, where $k$ was manually set, would be more insightful. $X^*$ deviated from the baseline method by $-0.07$, $-0.11$, and $+0.59$ in covering, PRI, and VI respectively. This deviation was due to approximating the input image by $m$ representatives then estimating the value of $k$ in two locations of the spectral clustering pipeline.

\begin{figure}
	\centering	
	\includegraphics[width=0.48\textwidth,height=20cm,keepaspectratio]{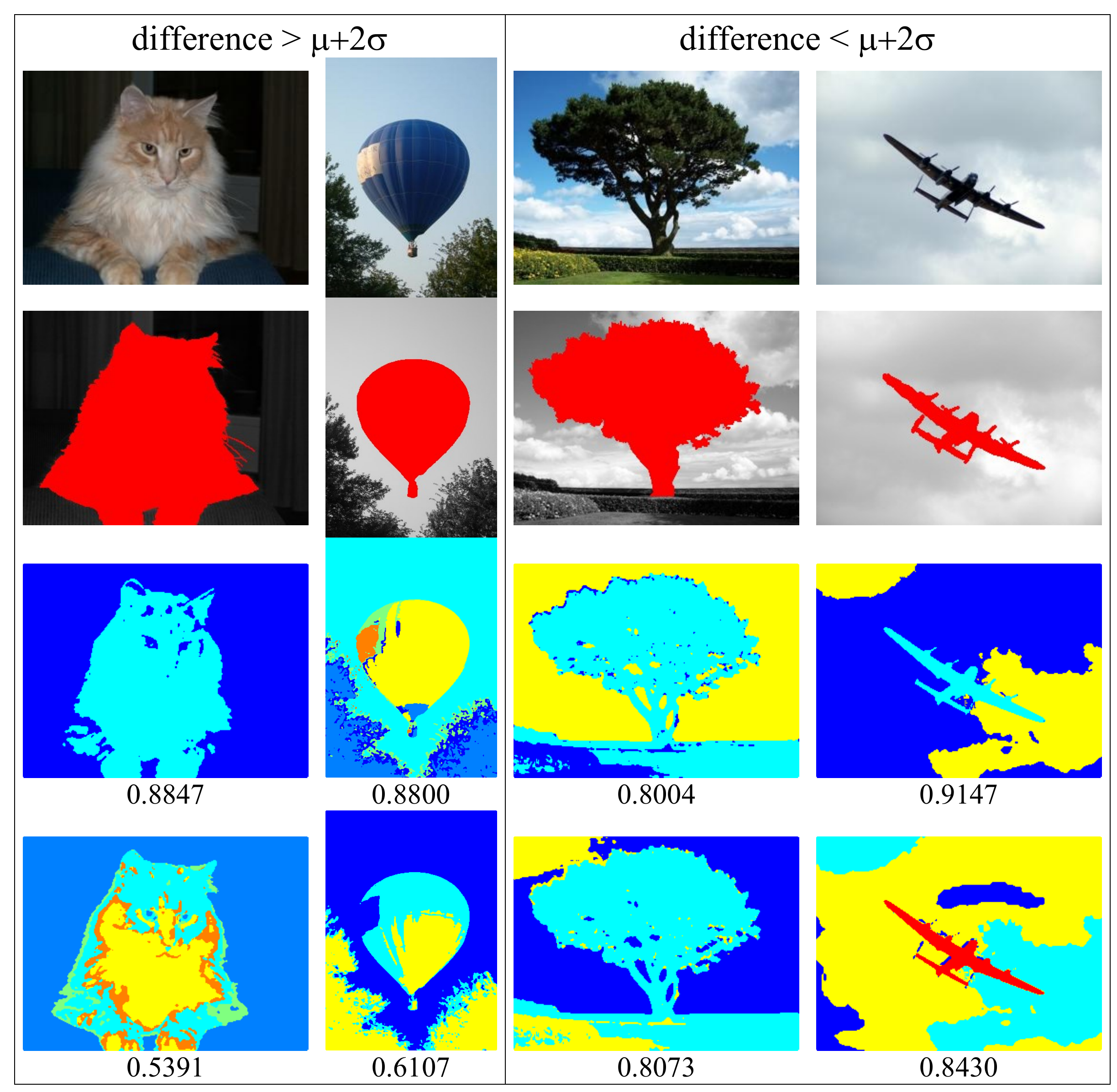}
	\caption{Examples for large and small differences between manual and auto settings. Rows ordered as: original, human segmentation, manual $k$, and auto $k$. The number beneath is the F-score (best viewed in color).}
	\label{Fig:ManualvsAuto}	
\end{figure}

We rerun Weizmann dataset with manual setting to compare against auto estimated $k$. The mean difference $\mu$ between manual and auto $k$ was 0.08 and the standard deviation $\sigma$ was 0.09. The difference was larger than $\mu + 2\sigma$ in 3 images. We suspect the human semantics incorporated in the ground truth cause these bad scores. For example, the image of the cat in Fig. \ref{Fig:ManualvsAuto} was segmented as per human perspective regardless of colors presence. Although the colors were segmented correctly by auto $k$, it was not the desired output by the human segmentation. On the other hand, when the human segmentation has limited colors, we could get competitive results as shown in Fig. \ref{Fig:ManualvsAuto}.

The final experiment was a comparison with some approximate spectral clustering methods. \cite{RN228} pulled out four images from BSDS500 to compare four approximate methods: multi-level low-rank approximate spectral clustering-original space (MLASC-O), Nystr{\"o}m-based spectral clustering (Nystr{\"o}m), INystr{\"o}m with $k$-means (INystr{\"o}m), and $k$-means approximate spectral clustering (KASP). We used the aforementioned methods as a baseline of our proposed approach, however $k$ was provided manually in these methods. As shown in Table \ref{Tab:Ex-Wang}, $X^*$ outperformed the competing methods in third and fourth images, with a clear advantage in the fourth one. For the first two images, $X^*$ deviated by $-0.0436$ and $-0.0330$ from the best performer in terms of segmentation covering.

\begin{table*}
	\centering
	\caption{Comparing the proposed method with the results reported in (\cite{RN228}). The number in parentheses indicates the deviation from the best score (in bold), and its color illustrates the change direction.}
	\label{Tab:Ex-Wang}
	\includegraphics[width=\textwidth,height=20cm,keepaspectratio]{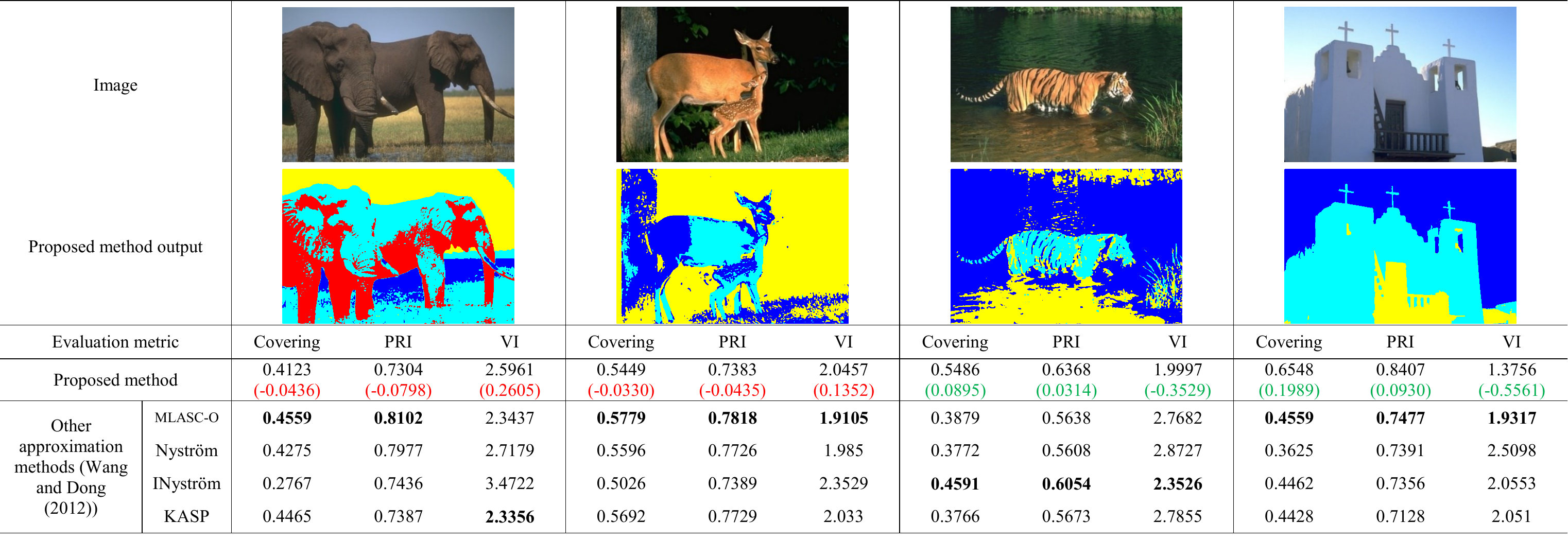}	
\end{table*}
\section{Conclusions}
\label{Conclusions}

Spectral clustering is an effective clustering tool that is able to detect non-convex shaped clusters. Despite its effectiveness, the computational demands of spectral clustering hold it back from being practically integrated into real applications. The efficient solution was to provide an approximation of input data. One factor was common among previous approximation techniques was the assumption of prior knowledge of the number of clusters $k$.

An automated estimation of $k$ alongside topology preserving approximation, were introduced in this work. We achieved that by setting two metrics, the former of which selects the embedding space dimensions $\mathbb{R}^{m \times k}$, while the latter detects the number of clusters in $\mathbb{R}^{m \times k}$. The first metric measure the relevance of an eigenvector $e_k$ based on its separation as well as possessing a small eigenvalue. Subsequently, the second cost function promotes the value of $k$ that separates the graph nodes better, and at the same time accumulates a small sum of eigenvalues. Experiments demonstrate that the proposed approach provides a competitive performance to the methods where $k$ was manually set.




\bibliographystyle{model2-names}
\bibliography{mybibfile}

\begin{thebibliography}{21}
\expandafter\ifx\csname natexlab\endcsname\relax\def\natexlab#1{#1}\fi
\providecommand{\url}[1]{\texttt{#1}}
\providecommand{\href}[2]{#2}
\providecommand{\path}[1]{#1}
\providecommand{\DOIprefix}{doi:}
\providecommand{\ArXivprefix}{arXiv:}
\providecommand{\URLprefix}{URL: }
\providecommand{\Pubmedprefix}{pmid:}
\providecommand{\doi}[1]{\href{http://dx.doi.org/#1}{\path{#1}}}
\providecommand{\Pubmed}[1]{\href{pmid:#1}{\path{#1}}}
\providecommand{\bibinfo}[2]{#2}
\ifx\xfnm\relax \def\xfnm[#1]{\unskip,\space#1}\fi
\bibitem[{Alpert et~al.(2007)Alpert, Galun, Basri and Brandt}]{RN221}
\bibinfo{author}{Alpert, S.}, \bibinfo{author}{Galun, M.},
  \bibinfo{author}{Basri, R.}, \bibinfo{author}{Brandt, A.},
  \bibinfo{year}{2007}.
\newblock \bibinfo{title}{Image segmentation by probabilistic bottom-up
  aggregation and cue integration}, in: \bibinfo{booktitle}{2007 IEEE
  Conference on Computer Vision and Pattern Recognition (CVPR)}, pp.
  \bibinfo{pages}{1--8}.
\newblock \DOIprefix\doi{10.1109/CVPR.2007.383017}.
\bibitem[{Arbelaez et~al.(2011)Arbelaez, Maire, Fowlkes and Malik}]{RN281}
\bibinfo{author}{Arbelaez, P.}, \bibinfo{author}{Maire, M.},
  \bibinfo{author}{Fowlkes, C.}, \bibinfo{author}{Malik, J.},
  \bibinfo{year}{2011}.
\newblock \bibinfo{title}{Contour detection and hierarchical image
  segmentation}.
\newblock \bibinfo{journal}{IEEE Transactions on Pattern Analysis and Machine
  Intelligence} \bibinfo{volume}{33}, \bibinfo{pages}{898--916}.
\newblock \DOIprefix\doi{10.1109/TPAMI.2010.161}.
\bibitem[{Bhatti et~al.(2018)Bhatti, Beck and Nedich}]{RN290}
\bibinfo{author}{Bhatti, S.}, \bibinfo{author}{Beck, C.},
  \bibinfo{author}{Nedich, A.}, \bibinfo{year}{2018}.
\newblock \bibinfo{title}{Data clustering and graph partitioning via simulated
  mixing}.
\newblock \bibinfo{journal}{IEEE Transactions on Network Science and
  Engineering} ,
  \bibinfo{pages}{1--1}\DOIprefix\doi{10.1109/TNSE.2018.2821598}.
\bibitem[{Cheung et~al.(2017)Cheung, Li, Peng and Chen}]{RN294}
\bibinfo{author}{Cheung, Y.}, \bibinfo{author}{Li, M.}, \bibinfo{author}{Peng,
  Q.}, \bibinfo{author}{Chen, C.L.P.}, \bibinfo{year}{2017}.
\newblock \bibinfo{title}{A cooperative learning-based clustering approach to
  lip segmentation without knowing segment number}.
\newblock \bibinfo{journal}{IEEE Transactions on Neural Networks and Learning
  Systems} \bibinfo{volume}{28}, \bibinfo{pages}{80--93}.
\newblock \DOIprefix\doi{10.1109/TNNLS.2015.2501547}.
\bibitem[{Cour et~al.(2005)Cour, Benezit and Shi}]{RN283}
\bibinfo{author}{Cour, T.}, \bibinfo{author}{Benezit, F.},
  \bibinfo{author}{Shi, J.}, \bibinfo{year}{2005}.
\newblock \bibinfo{title}{Spectral segmentation with multiscale graph
  decomposition}.
\newblock \bibinfo{journal}{2005 IEEE Conference on Computer Vision and Pattern
  Recognition (CVPR)} \bibinfo{volume}{2}, \bibinfo{pages}{1124--1131 vol. 2}.
\newblock \DOIprefix\doi{10.1109/CVPR.2005.332}.
\bibitem[{Fritzke(1995)}]{RN270}
\bibinfo{author}{Fritzke, B.}, \bibinfo{year}{1995}.
\newblock \bibinfo{title}{A growing neural gas network learns topologies}.
\newblock \bibinfo{journal}{Advances in Neural Information Processing Systems}
  , \bibinfo{pages}{625--632}.
\bibitem[{Huiskes and Lew(2008)}]{RN249}
\bibinfo{author}{Huiskes, M.J.}, \bibinfo{author}{Lew, M.S.},
  \bibinfo{year}{2008}.
\newblock \bibinfo{title}{The mir flickr retrieval evaluation}.
\newblock \bibinfo{journal}{Proceedings of the 1st ACM international conference
  on Multimedia information retrieval} , \bibinfo{pages}{39--43}.
\bibitem[{Li et~al.(2017)Li, Fan, Sun, Li, Chen and Liu}]{RN293}
\bibinfo{author}{Li, Q.}, \bibinfo{author}{Fan, H.}, \bibinfo{author}{Sun, W.},
  \bibinfo{author}{Li, J.}, \bibinfo{author}{Chen, L.}, \bibinfo{author}{Liu,
  Z.}, \bibinfo{year}{2017}.
\newblock \bibinfo{title}{Fingerprints in the air: Unique identification of
  wireless devices using rf rss fingerprints}.
\newblock \bibinfo{journal}{IEEE Sensors Journal} \bibinfo{volume}{17},
  \bibinfo{pages}{3568--3579}.
\newblock \DOIprefix\doi{10.1109/JSEN.2017.2685564}.
\bibitem[{Liu et~al.(2013)Liu, Lin, Yan, Sun, Yu and Ma}]{RN168}
\bibinfo{author}{Liu, G.}, \bibinfo{author}{Lin, Z.}, \bibinfo{author}{Yan,
  S.}, \bibinfo{author}{Sun, J.}, \bibinfo{author}{Yu, Y.},
  \bibinfo{author}{Ma, Y.}, \bibinfo{year}{2013}.
\newblock \bibinfo{title}{Robust recovery of subspace structures by low-rank
  representation}.
\newblock \bibinfo{journal}{IEEE Transactions on Pattern Analysis and Machine
  Intelligence} \bibinfo{volume}{35}, \bibinfo{pages}{171--184}.
\newblock \DOIprefix\doi{10.1109/TPAMI.2012.88}.
\bibitem[{von Luxburg(2007)}]{RN234}
\bibinfo{author}{von Luxburg, U.}, \bibinfo{year}{2007}.
\newblock \bibinfo{title}{A tutorial on spectral clustering}.
\newblock \bibinfo{journal}{Statistics and Computing} \bibinfo{volume}{17},
  \bibinfo{pages}{395--416}.
\newblock \DOIprefix\doi{10.1007/s11222-007-9033-z}.
\bibitem[{Martinetz(1993)}]{RN269}
\bibinfo{author}{Martinetz, T.}, \bibinfo{year}{1993}.
\newblock \bibinfo{title}{Competitive hebbian learning rule forms perfectly
  topology preserving maps}.
\newblock \bibinfo{journal}{Artificial Neural Networks and Machine Learning -
  {ICANN} 1993} , \bibinfo{pages}{427--434}\URLprefix
  \url{https://doi.org/10.1007/978-1-4471-2063-6_104},
  \DOIprefix\doi{10.1007/978-1-4471-2063-6_104}.
\bibitem[{Ng et~al.(2002)Ng, Jordan and Weiss}]{RN233}
\bibinfo{author}{Ng, A.Y.}, \bibinfo{author}{Jordan, M.I.},
  \bibinfo{author}{Weiss, Y.}, \bibinfo{year}{2002}.
\newblock \bibinfo{title}{On spectral clustering: Analysis and an algorithm}.
\newblock \bibinfo{journal}{Advances in Neural Information Processing Systems}
  .
\bibitem[{Shi and Malik(2000)}]{RN261}
\bibinfo{author}{Shi, J.}, \bibinfo{author}{Malik, J.}, \bibinfo{year}{2000}.
\newblock \bibinfo{title}{Normalized cuts and image segmentation}.
\newblock \bibinfo{journal}{IEEE Transactions on Pattern Analysis and Machine
  Intelligence} \bibinfo{volume}{22}, \bibinfo{pages}{888--905}.
\newblock \DOIprefix\doi{10.1109/34.868688}.
\bibitem[{Tyuryukanov et~al.(2018)Tyuryukanov, Popov, Meijden and
  Terzija}]{RN288}
\bibinfo{author}{Tyuryukanov, I.}, \bibinfo{author}{Popov, M.},
  \bibinfo{author}{Meijden, M.v.d.}, \bibinfo{author}{Terzija, V.},
  \bibinfo{year}{2018}.
\newblock \bibinfo{title}{Discovering clusters in power networks from
  orthogonal structure of spectral embedding}.
\newblock \bibinfo{journal}{IEEE Transactions on Power Systems}
  \DOIprefix\doi{10.1109/TPWRS.2018.2854962}.
\bibitem[{Vora and Raman(2018)}]{RN286}
\bibinfo{author}{Vora, A.}, \bibinfo{author}{Raman, S.}, \bibinfo{year}{2018}.
\newblock \bibinfo{title}{Iterative spectral clustering for unsupervised object
  localization}.
\newblock \bibinfo{journal}{Pattern Recognition Letters} \bibinfo{volume}{106},
  \bibinfo{pages}{27--32}.
\newblock \URLprefix
  \url{http://www.sciencedirect.com/science/article/pii/S0167865518300473},
  \DOIprefix\doi{https://doi.org/10.1016/j.patrec.2018.02.012}.
\bibitem[{Wang and Dong(2012)}]{RN228}
\bibinfo{author}{Wang, L.}, \bibinfo{author}{Dong, M.}, \bibinfo{year}{2012}.
\newblock \bibinfo{title}{Multi-level low-rank approximation-based spectral
  clustering for image segmentation}.
\newblock \bibinfo{journal}{Pattern Recognition Letters} \bibinfo{volume}{33},
  \bibinfo{pages}{2206--2215}.
\newblock \DOIprefix\doi{http://dx.doi.org/10.1016/j.patrec.2012.07.024}.
\bibitem[{Wang et~al.(2017)Wang, Lin and Wang}]{RN292}
\bibinfo{author}{Wang, T.S.}, \bibinfo{author}{Lin, H.T.},
  \bibinfo{author}{Wang, P.}, \bibinfo{year}{2017}.
\newblock \bibinfo{title}{Weighted-spectral clustering algorithm for detecting
  community structures in complex networks}.
\newblock \bibinfo{journal}{Artificial Intelligence Review}
  \bibinfo{volume}{47}, \bibinfo{pages}{463--483}.
\newblock \URLprefix \url{https://doi.org/10.1007/s10462-016-9488-4},
  \DOIprefix\doi{10.1007/s10462-016-9488-4}.
\bibitem[{Xiang and Gong(2008)}]{RN265}
\bibinfo{author}{Xiang, T.}, \bibinfo{author}{Gong, S.}, \bibinfo{year}{2008}.
\newblock \bibinfo{title}{Spectral clustering with eigenvector selection}.
\newblock \bibinfo{journal}{Pattern Recognition} \bibinfo{volume}{41},
  \bibinfo{pages}{1012--1029}.
\newblock \URLprefix
  \url{http://www.sciencedirect.com/science/article/pii/S0031320307003688},
  \DOIprefix\doi{https://doi.org/10.1016/j.patcog.2007.07.023}.
\bibitem[{Yan et~al.(2009)Yan, Huang and Jordan}]{RN232}
\bibinfo{author}{Yan, D.}, \bibinfo{author}{Huang, L.},
  \bibinfo{author}{Jordan, M.I.}, \bibinfo{year}{2009}.
\newblock \bibinfo{title}{Fast approximate spectral clustering}.
\newblock \bibinfo{journal}{Proceedings of the 15th ACM SIGKDD international
  conference on Knowledge discovery and data mining} ,
  \bibinfo{pages}{907--916}.
\bibitem[{Zelnik-Manor and Perona(2005)}]{RN237}
\bibinfo{author}{Zelnik-Manor, L.}, \bibinfo{author}{Perona, P.},
  \bibinfo{year}{2005}.
\newblock \bibinfo{title}{Self-tuning spectral clustering}.
\newblock \bibinfo{journal}{Advances in Neural Information Processing Systems}
  , \bibinfo{pages}{1601--1608}.
\bibitem[{Zhao et~al.(2010)Zhao, Jiao, Liu, Gao and Gong}]{RN264}
\bibinfo{author}{Zhao, F.}, \bibinfo{author}{Jiao, L.}, \bibinfo{author}{Liu,
  H.}, \bibinfo{author}{Gao, X.}, \bibinfo{author}{Gong, M.},
  \bibinfo{year}{2010}.
\newblock \bibinfo{title}{Spectral clustering with eigenvector selection based
  on entropy ranking}.
\newblock \bibinfo{journal}{Neurocomputing} \bibinfo{volume}{73},
  \bibinfo{pages}{1704--1717}.
\newblock \URLprefix
  \url{http://www.sciencedirect.com/science/article/pii/S0925231210001311},
  \DOIprefix\doi{https://doi.org/10.1016/j.neucom.2009.12.029}.

\end{thebibliography}

\end{document}